\documentclass[review]{elsarticle}

\usepackage{lineno}
\usepackage[normalem]{ulem}
\let\linenumbers\nolinenumbers\nolinenumbers

\journal{Neural Networks}

\usepackage[colorlinks=true]{hyperref}
\usepackage{caption}
\usepackage{subcaption}
\usepackage{color}
\usepackage{amsmath}
\usepackage{multirow}
\usepackage[table,xcdraw]{xcolor}

\newcommand{\methodname}{DMBN}

\definecolor{erhcol}{rgb}{0.5, 0.7, 0.0}
\newcommand{\erhan}[1]{\textcolor{black}{#1}} %brown
\newcommand{\erhann}[1]{\textcolor{black}{#1}} %erhcol
\newcommand{\erhanadd}[1]{\textcolor{black}{#1}} %black
\newcommand{\emree}[1]{\textcolor{black}{#1}} %blue

\begin{document}

\begin{frontmatter}

\title{Imitation and Mirror Systems in Robots through \\
Deep Modality Blending Networks}
%End-to-end, high-dimensional, flexible conditioned,                                               Any time/duration/modality conditioned Imagination of multi-modal trajectories and mirror neuron emergence
%An end-to-end temporal multi-modal deep learning architecture addressing imitation and mirror systems %in biological systems
%ALT1. Deep Modality Blending for Action Recognition and Imitation
% action recognition yanlış anlaşılabilir, direk videodan falan actionı segment eden işler de action recog. diyorlar.
%ALT2. Deep Modality Blending Network Addressing Imitation and Mirror Systems  \\
%ALT3. Deep Modality Blending Network for Egocentric and Mirror Imitation


\author[boun]{M. Yunus Seker\corref{mycorrespondingauthor}}
\ead{yunus.seker1@boun.edu.tr}
\cortext[mycorrespondingauthor]{Corresponding author}

\author[boun]{Alper Ahmetoglu}
% \ead{alper.ahmetoglu@boun.edu.tr}

\author[tokyo]{Yukie Nagai}
% \ead{nagai.yukie@mail.u-tokyo.ac.jp }

\author[osaka]{Minoru Asada}
% \ead{asada@otri.osaka-u.ac.jp}

\author[osaka,ozu]{Erhan Oztop}
% \ead{erhan.oztop@otri.osaka-u.ac.jp}

\author[boun]{Emre Ugur}
% \ead{emre.ugur@boun.edu.tr}

\address[boun]{Bogazici University, Bebek, Istanbul, 34342, Turkey}
\address[osaka]{Osaka University, Suita, Osaka, Japan}
\address[ozu]{Ozyegin University, Istanbul, Turkey}
\address[tokyo]{The University of Tokyo, Bunkyo-ku, Tokyo, Japan}

\begin{abstract}
%Imitation learning is one popular method to equip robots with desired control functions to perform a task by providing or self-exploring a set of demonstrations in the robot's proprioceptive space. Although there are imitation learning frameworks that allow learning a policy from multiple modalities, most of these frameworks treat multiple modalities only as an extra information channel. 

Learning to interact with the environment not only empowers the agent with manipulation capability but also generates information to facilitate building of action understanding and imitation capabilities. This seems to be a strategy adopted by biological systems,  in particular primates, as evidenced by the existence of mirror neurons that seem to be involved in multi-modal action understanding. How to benefit from the interaction experience of the robots to enable understanding actions and goals of other agents is still a challenging question. In this study, we propose a novel method, deep modality blending networks (DMBN), that creates a common latent space from multi-modal experience of a robot by blending multi-modal signals with a stochastic weighting mechanism. We show for the first time that deep learning, when combined with a novel modality blending scheme, can facilitate action recognition and produce structures to sustain anatomical and effect-based imitation capabilities. Our proposed system, which is based on conditional neural processes, can be conditioned on any desired sensory/motor value at any time-step, and can generate a complete multi-modal trajectory consistent with the desired conditioning in one-shot by querying the network for all the sampled time points in parallel avoiding accumulation of prediction errors. Based on simulation experiments with an arm-gripper robot and an RGB camera, we showed that DMBN could make accurate predictions about any missing modality (camera or joint angles) given the available ones outperforming recent multimodal variational autoencoder models in terms of long-horizon high-dimensional trajectory predictions. We further showed that given desired images from different perspectives, i.e. images generated by the observation of other robots placed on different sides of the table, our system could generate image and joint angle sequences that correspond to either anatomical or effect based imitation behavior. To achieve this mirror-like behavior our system does not perform a pixel-based template matching but rather benefits from and relies on the common latent space constructed by using both joint and image modalities, as shown by additional experiments. Overall, the proposed DMBN architecture not only serves as a  computational model for sustaining mirror neuron-like capabilities, but also stands as a powerful machine learning architecture for high-dimensional multi-modal temporal data with robust retrieval capabilities operating with partial information in one or multiple modalities.

\end{abstract}

%\begin{keyword}
%\end{keyword}

\end{frontmatter}

\linenumbers

\section{Introduction}

With appropriate and sufficient amount of  data, a range of sensorimotor learning tasks encountered by robots and biological systems can be solved by deep learning. However, unlike the abundance of data for image recognition and language modeling, robots and biological systems often need to harvest data themselves by either using self-exploration based learning or by observing the relevant behaviors of other agents. These two alternatives are studied in robotics and machine learning under the general titles of  Reinforcement Learning (RL) \cite{Sutton2018} and Learning from Demonstration (LfD)\cite{Argall2009}.
%Although approaches involving RL and LfD are on the rise \cite{Argall2009,akbulut2020acnmp}, 
Although the use of self-observation during self-executed actions is common for forming a reward signal in RL, it is not well addressed how to benefit the agent in a cognitive developmental sense, for example, for recognizing actions of others or forming a general imitation capacity. Learning to interact with the environment not only empowers the agent with manipulation capability but also generates information to facilitate the building of action understanding and imitation capabilities. This seems to be a strategy adopted by biological systems, in particular primates, as evidenced by the existence of mirror neurons \cite{Pellegrino1992, Rizzolatti_etal_1996} in the ventral premotor cortex of those animals, which encode actions in a multi-modal fashion \cite{Kohler_etal_2002}. For example, there are mirror neurons that become active when the animal breaks a peanut, observes an experimenter do the same act or hears the sound of peanut cracking \cite{Keysers_etal_2003}. With such a system, sensed actions are mapped to one's own motor representation; and thus can bootstrap imitation, by for example, understanding the parts of an observed act in terms of the existing `action vocabulary' of the animal, which can be
%chained to reproduce the recognized action segments yielding novel action imitation capability. 
reproduced in sequence yielding novel action imitation capability. 
Although, it is not clear whether mirror neurons play a role in imitation, as their exact function and mechanism are far from clear, computational modeling may help produce insights towards understanding them \cite{OztopKawatoArbib2013}. Therefore,
from a scientific and also a technological point of view, it is desirable to develop a neural multi-modal action representation system that can learn/store actions and recall them from partial information that might be transformed as in the case of action observation from different perspectives. In fact, there exist a range of computational models related to mirror neurons and their function in the literature (\cite{Oztop2002,Bonaiuto2007,Bonaiuto2010, Copete2016, TaniItoSugita2004, DemirisJohnson2003}) that have leveraged our understanding by creating hypotheses to be tested. Now, the time is ripe for a less constrained, end-to-end and more powerful multi-modal action representation mechanism for obtaining better insights.
%in the midst of the deep learning era. 
In particular, the existing multi-modal action representation schemes based on self-observation either fall short of providing robust recognition and imitation capability or rely on feature engineering.  

In this study, we improve the state of the art in multi-modal action representation by showing for the first time that deep learning, when combined with a novel modality blending scheme, can facilitate feature-engineering-free action recognition and basic imitation capabilities under perspective changes with only  partial information. Moreover, the modality blending scheme produces latent representations that can sustain both anatomical and effect-based imitation capabilities. We call the developed multi-modal action representation architecture as a Deep Modality Blending Network (DMBN).

\begin{figure}[t]
    \centering
    \includegraphics[width=\linewidth]{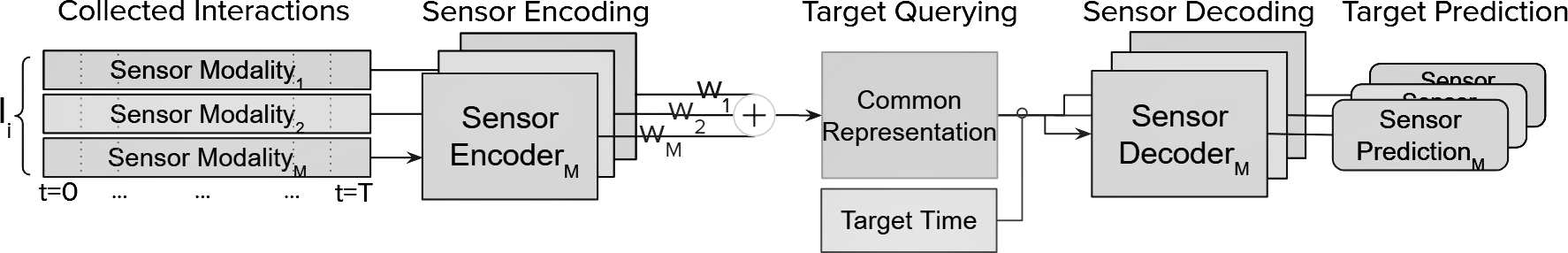}
    \caption{General architecture of a Deep Modality Blending Network}
    \label{fig:generalfig}
\end{figure}
DMBN connects multiple modalities by blending them as random mixtures of modality-specific latent representations to form a common latent representation for seamless transfer from one modality to another (see Figure~\ref{fig:generalfig}).  The DMBN architecture  follows an encoder-decoder structure where each modality is summarized by its corresponding encoder network, processing the sensorimotor data into a compact latent representation. While learning, not only these latent representations are formed but they are blended together into a common representation through stochastic mixture weights. After learning, using the common representation, each decoder network can predict the corresponding modality for an arbitrary desired time-step,  effectively generating outcome predictions as temporal sequences for all the modalities. 
\erhann{In this sense, the common latent layer in our network encodes representation of the complete multi-modal trajectories rather than encoding modalities in particular time steps. This feature sets our system apart from its competitors \cite{zambelli2020multimodal,Copete2016} and give it a big advantage. To be concrete, our system can be conditioned on any desired sensory/motor value at any time-step, and can generate a complete multi-modal trajectory consistent with the desired conditioning in one-shot by querying the network for all the sampled time points in parallel. This one-shot full trajectory decoding ability makes our system very accurate as it does not suffer from the error accumulation faced by systems that need to chain next-state predictions in order to generate full trajectories. }

%\emree{Built on top of a recent neural system, namely Conditional Neural Processes, the common latent layer in our network encodes representation of the complete multi-modal trajectories rather than encoding modalities in particular time steps. Our system, different from its competitors \cite{zambelli2020multimodal,Copete2016}, can be conditioned with any desired sensory/motor value at any time-step, and can generate the complete multi-modal trajectory consistent with the desired values in one-shot by querying the network for all time points in parallel. Therefore, our system is very flexible in the sense that it can generate full trajectories given desired values at any desired time-points, and is accurate in the sense that it does not suffer from the error accumulation faced by systems that need to chain next-state predictions in order to generate full trajectories.}

\emree{To demonstrate the efficacy of the proposed DMBN architecture, we implemented it in a simulated manipulation setup. In this setup, an object was placed in the middle of a table, and an arm-gripper robot was set to execute grasp and push actions on it with different approach directions. The robot observed the consequences of its actions using an RGB camera from a fixed perspective, and learned the generated multi-modal sensory (visual and proprioceptive) signals as sensory trajectory distributions through the proposed DMBN architecture. After learning, }
%\emree{Our system was realized in a simulated manipulation system. Given an object in the middle of the table, an arm-gripper robot was programmed to execute grasp and push actions on this object with different approach directions, observed the consequences of its actions using an RGB camera from a fixed perspective, and learned multi-modal (camera and joint) trajectory distributions through the previously described encoder-decoder structure with through our stochastic blending and conditional processing approach. After learning, }
\emree{
\begin{itemize}
    \item Given desired images at any time point (such as images of objects lifted or pushed away), our system can find the joint trajectories that are required to generate changes in the environment to observe these images;
    \item Given joint angles at any time point(s), our system can generate the sequence of images that are expected to be observed during the execution of the action that is consistent with given angles;
    \item Given desired images from different perspectives, i.e. images generated by the observation of other robots placed on different sides of the table, our system can generate image and joint angle sequences that correspond to valid actions of the robot;
    \item Those valid actions, intriguingly correspond to either anatomical or effect based imitation  behavior.
    %\item given desired images from different perspectives, i.e. images generated by \erhann{the observation of other} robots placed on different sides of the table, our system can generate the corresponding images from its own perspective along with the corresponding joint angles.
\end{itemize}
%Interestingly, given desired images from different perspectives, our system can generates mirror-like or goal-emulating behavior. 
%\erhann{Interestingly, given desired images from different perspectives, our system can generate anatomical  or effect based imitation  behavior.} 
To clarify the last bullet above it would be useful to consider an example behavior observed in our simulations.
Given an image that shows the snapshot of another robot on the other side of the table pulling the object to itself, our system can generate the sequence of images where its own gripper pulls the object towards itself (anatomical imitation behavior) or pushes the object towards the other side of the table (effect based imitation or goal-emulation behavior) depending on the visual cues available to the robot. 
In our analysis, we show that the prediction capability of the proposed DMBN system does not simply perform a pixel-based template matching} but rather benefits from and relies on the common latent space constructed by using both joint and image modalities.
%In our analysis, we showed that the prediction capability of our system, when queried from different perspectives, does not utilize pixel-based-only matching but rather benefits from and relies on the common latent space constructed using both joint and image modalities.
In addition to other interesting results, our experiments clearly show that our system outperforms a recent multimodal variational autoencoder model \cite{zambelli2020multimodal} in reconstructing long-horizon  high-dimensional trajectories.

%Our main contribution in this paper is to propose the multi-modality blending mechanism which stands in the middle of our architecture. Since the blending is realized by random weight mixtures, after some training, latent outputs of the modalities converge to common representations in order to reduce the stochasticity of the mixed representation space, and to optimize the neural network accuracy. At the end of the learning, encoders of different modalities output nearly the same representation for the same state, thus, making it is easy to transfer one modality to another at the test time. This transferability allows the system to predict sensorimotor values of missing modalities given any other available one. For example, given a visual observation, the agent can infer the action values (i.e. joint angles) of a manipulation action, or vice versa. Our experimental results show that as our network learns the relationship between multiple modalities, it \erhan{builds} advanced cognitive \erhan{mechanisms} such as \erhan{anatomical and effect based imitation ability when the network is subject to observation of actions that were not experienced during learning}.

%\erhcom{MNs do not have two type imitation as an essential feature!:}
%\erhcom{So here what exactly do you want to say? (Rewrite without MN reference):
%Furthermore our experiments show that learning multiple modalities with modality blending play a significant role for the development of such mirror neuron systems.}

The outline of this paper is as follows: in Section \ref{sec:related}, we review the related work, in particular LfD systems as DMBN builds upon one such system and the competing multi-modal action representations. In Section \ref{sec:method}, we describe our proposed method in detail. We explain our experiment setup in Section \ref{sec:expsetup} and give experimental results in Section \ref{sec:expresult}. Finally, we give a conclusion in Section \ref{sec:conc}.

\section{Related Work}
\label{sec:related}
% ... . Although RL has been shown to be very effective on various tasks \cite{mnih2013playing,silver2016mastering,vinyals2017starcraft,andrychowicz2020learning}, in some cases it requires lots of trial and errors which might not be always feasible in the real-world setups either due to safety standards or time restrictions. For this reason, i
Imitation learning, or learning from demonstration (LfD) \cite{Argall2009}, has been a popular research topic in robotic learning \cite{Pastor2011,neumann2018,Asfour2008,Pastor2009,Ben2012,Muhlig2012}. Various LfD methods has been proposed based on dynamic systems and statistical modeling \cite{Schaal2006, TamimTPDMP17, Calinon2016, kernelizedmp}, where the parameters in the environment can be learned with Locally Weighted Regression \cite{Atkeson1997,Ude2010task,Ude17} and Locally Weighted Projection Regression \cite{Vijayakumar2000}. Gaussian Mixture Models \cite{Calinon2009,LeeTPDMP17} and Hidden Markov Models \cite{Lee2011,Chu2013,Girgin2018,ugur19} are also frequently used to learn the motion distributions from multiple demonstrations. More recently, deep neural networks also started to be used in imitation learning in order to make it possible to learn movement primitives from complex high-dimensional data \cite{deepimitation, pervez2017, Morimoto2018deepDMP,Droniou2015}. %\cite{Droniou2015} learns multi-modal models from various sensory data such as pressure, temperature, proprioception, and speech; however these models were used to distinguish the sensory data and was not affecting the execution of the actions. 
In our earlier work, we proposed Conditional Neural Movement Primitives (CNMPs) \cite{sekerconditional} as an end-to-end deep LfD architecture that can learn temporal sensorimotor distributions of complex manipulation skills. The  DMBN architecture developed in the current study, builds upon CNMPs by introducing a novel mechanism for  modality blending to learn a common latent representation that allows cross-modal temporal prediction with partial information.
%where Akbulut et al. \cite{akbulut2020acnmp,Akbulut-2021-ICRA} added a reinforcement learning mechanism %These models were successful on learning complex high dimensional data with non-linear relations, however they are often used to learn a single modality.

Several works studied the emergence of the mirror neuron system (MNS) in the context of multi-modal sensor fusion. Nagai et al. \cite{asada2011} proposed a computational model for the early development of the MNS. In this model, the robot cannot make self-other discrimination in the early stages due to the immature visual system. As the visual system develops, the robot starts to discriminate between itself and others, yet, still retains information regarding early experiences, producing the MNS as a by-product. Noda et al. \cite{ogata2014} used time-delay neural networks \cite{waibel1989phoneme} as autoencoders to fuse multiple modalities and reconstruct the missing ones given others. 

Copete et al. \cite{yukie2016} also used a similar autoencoder architecture in a predictive learning context so to imagine the action of others. Jung et al. \cite{tani2019} proposed a top-down visual attention system to address the long-term visual prediction problem. In this system, the visual stream is divided into dorsal and ventral streams to decompose the difficulty of the problem into two sub-problems. These two streams are then merged for the visual prediction with the help of an external visuospatial memory which holds long-term visuospatial information. On the other hand, we provide a more holistic approach where there are only different submodules for different modalities. Our experiments show that DMBNs can output very accurate visual signals conditioned only on a single visual frame without any memory module. Among these studies, the learning problems considered in the work of Zambelli et al. \cite{zambelli2020multimodal} is well aligned with our study. They proposed a multimodal variational autoencoder (MVAE) \cite{suzuki2016joint,wu2018multimodal} to fuse the sensorimotor information of an iCub humanoid robot for prediction and control. They showed that by training MVAE as a denoising autoencoder \cite{vincent2008extracting}, MVAE can predict the future sensorimotor states, reconstruct the missing modalities, and imitate based on human action observation. As MVAE is not a recurrent architecture, the temporal information should be explicitly stated in the input. To be concrete, in the training phase, the sensorimotor information at time $t$ and $t+1$ were combined and given as input to the MVAE for reconstruction. Here, some sensorimotor information at time $t+1$ was randomly masked with -2 (as in a denoising autoencoder) in order to train the network to reconstruct  the future timestep even if it was partially missing.

In the testing phase for future state predictions, states at $t+1$ were filled with mask values -2. Further steps could be predicted by feeding the output of the MVAE to the input. However, the error at one step cascades in the feedback loop as in RNNs. Therefore, the prediction power decreases as the trajectory horizon increases. This is not the case in our proposed model as DMBNs make temporal prediction in one-shot without requiring feeding back of the output as input. To concretely state, our work differs from the previous works in terms of modality fusion strategy and architecture: (1) we take a stochastic mixture of modalities to force the formation of a more regularized representation, and (2) we learn individual modalities and their mixture as long range dependencies via CNPs \cite{CNP}, which allow arbitrary future and past temporal predictions. These key differences yield not only a more robust and better performing multi-modal action representation system, but also give raise to interesting generalization abilities as shown with the experiments presented in the Results section.

\section{Method}
\label{sec:method}

\begin{figure*}[h]
    \centering
    \includegraphics[width=\linewidth]{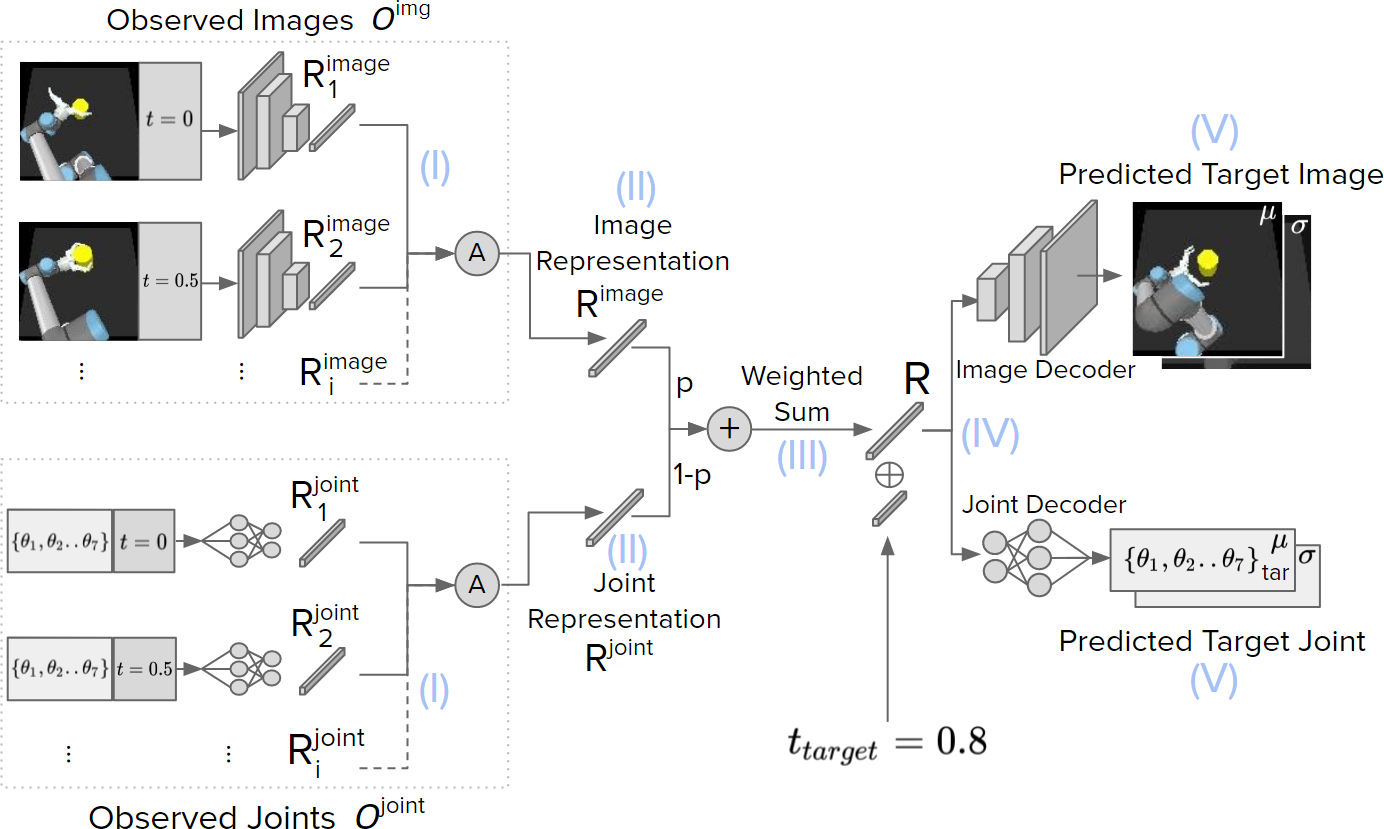}
    \caption{Proposed framework for given $visual$ and $joint$ modalities. Image and joint observations are turned into their latent representations separately to be used to predict the image and joint positions given at another target time step.}
    \label{fig:mainfig}
\end{figure*}

In this work, we propose Deep Modality Blending Networks (\methodname s), that can learn and produce sensorimotor signals by forming and exploiting multi-modal representations acquired in a latent space. Assume $M$ = \{visual, proprioception, sound, haptic ...\} corresponds to sensorimotor signals from multiple modalities that an agent collects through self-observation. The agent interacts with the environment using a variety of actions to leverage the information produced by the embodied interaction of the agent with the environment. In the current implementation, the action and action parameters are sampled from a  predefined action repertoire. During every interaction, the sensorimotor values are recorded at each time-step. The multi-sensorimotor interaction data set is defined as $I$, and the $i$th interaction is described as $I_i = \{(t,S^M_t)\}_{t=0}^{T}$, where $t$ is time and $S^M_t$ is the sensorimotor state collection for the given time-step. $S^M_t$ consists of multiple sensorimotor data, $S^M_t = [S^{visual}_t,S^{joint}_t,S^{sound}_t,S^{haptic}_t, ...]$, where each member holds the corresponding state values of the sensorimotor modalities for the time-step $t$. Figure \ref{fig:mainfig} shows the architecture of our model where the modalities in the system correspond to the $visual$ and $proprioceptive$ domains. These two domains are chosen specifically in order to show that our system can learn in an end-to-end fashion with both high (image) and low (joint) dimensional data and make more accurate target predictions on a long horizon compared to the sequential prediction models. In theory, all types of sensorimotor data can be included in the system with our formulation. 

The aim of \methodname~is to predict a conditional output distribution for a target query given a desired set of observation samples. At the beginning of each training iteration, an interaction $I_d$ is selected randomly from the data set $I$. From this selected interaction, $n$ data points of $(t,S^M_t)$, are randomly sampled as observations. Here, $n$ is a changing number for each training iteration that is bounded by $[1,obs_{max}]$ where $obs_{max}$ is a hyper-parameter that decides the maximum number of sampled observations in the training. We define this sampled observation set as $O^M = \{(t_i,S^M_{t_i})\}_i^{obs_{max}}$ where $(t_i,S^M_{t_i}) \in I_d$. On the left side of the Figure \ref{fig:mainfig}.(I), example sampled observations $O^{image}$ and $O^{joint}$ are shown for the image and the joint domains. Besides $O^M$, a target tuple $(t_{target},S^M_{t_{target}})$ is also sampled from the same selected interaction $I_d$. The purpose of a training iteration is to learn distributions on $t_{target}$ for all modalities in the system, based on the observation set $O^M$.

Our aim is to merge the observations of all the modality signals in a single latent space to allow information sharing for a higher quality prediction. In order to achieve this, the observations of each modality, $O^m$, are first turned into their latent representations $R^m_i$. For every modality $m$ and every observation, latent states are calculated by the following equation:
\begin{align}
  R^m_i = E^m((t_i,S^m_{t_i}) \mid \theta^m) && (t_i,S^m_{t_i}) \in O^m, m \in M
\end{align}
where $E^m$ is a deep encoder for the modality $m$ with weights $\theta^m$, and $R^m_i$ is the latent states of its $i$th observation. Figure \ref{fig:mainfig}.(I) shows the encoded representations, $R^{image}_i$ and $R^{joint}_i$, for each observation. After generating these representations, an averaged representation of each modality is calculated by: 
\begin{align}
  R^m = \frac{1}{n} \sum_i^{n} R^m_i && m \in M
  \label{eq:avg_latents}
\end{align}
where $n$ is the size of the observations of this training iteration. $R^{image}$ and $R^{joint}$ in Figure 2(II) hold general knowledge about their modalities, and our aim is to use these representations in a shared latent space to allow information sharing between all the modalities. To achieve that, a multi-modal general representation $R$ that integrates all the modalities is constructed by calculating a normalized weighted average:
\erhanadd{
\begin{align}
R = \frac{\sum_m^M p^m R^m w^m}{\sum_m^M p^m  w^m}
\label{Eq_normalization}
\end{align}
where $w^M=[w^{image},w^{joint},w^{m},...]$ is a vector representing the \textit{weight} or \textit{availability} of the individual modalities with $0\le w^M\le 1$ and $w^M \neq 0$, which could be used to model cases where one modality is more reliable than the other. On the other hand, modality blending during training is achieved through the random variables $0 \le p^m \le 1$  that is sampled at every iteration, and obey the constraint $\sum p^m=1$. Note that  to avoid  $\sum_m^M p^m w^m$ ever becoming zero (See Eq~\ref{Eq_normalization}), we may require $p^m>0$; but this is not an issue in practice. }
%\begin{align}
%R = \frac{\sum_m^M R^m * w^m}{\sum_m^M w^m}
%\end{align}
%where $w^M=[w^{image},w^{joint},w^{m},...]$ is a vector that represents the \textit{availability} of the modality $m$ for that training iteration. Although every modality of the training data might be fully available, we train our model in a way that each value in $w^M$ is in the range $[0,1]$ and randomly change in every training iteration to make the framework robust to the changes at test time. 
This follows the same intuition with dropout \cite{srivastava2014dropout}; randomly dropping modalities forces the model to learn compact representations that can compensate for missing information. Figure \ref{fig:mainfig}.(III) shows this process as a two-modality setup where 
%the values of $w^{image}$ and $w^{joint}$ are $p$ and $1-p$ respectively. Here, $p$ is a random number bounded by $[0,1]$ that changes every training iteration. 
\erhanadd{ $w^{image} = w^{joint} = 0.5$ and  $p^{image} = p$ and  $p^{joint}=1-p$ where $p$ is sampled uniformly from [0,1].}
Note that the dimension of each $R^m$ should be the same in order to perform summation operation between vectors, so in the first place, all the encoders must be designed to produce the latent states with the same dimensions. Once all observations are merged into one general representation, this information can be used to infer target distributions on $t_{target}$ for all the modalities as:
\begin{align}
(\mu^m_{t_{target}},\sigma^m_{t_{target}}) = Q^m((R, t_{target}) \mid \phi^m) && m \in M 
\end{align}
where $Q^m$ is a deep decoder network with weights $\phi^m$ that produces a distribution that consists of a mean $\mu_{t_{target}}^m$ and variance $\sigma_{t_{target}}^m$ for the modality $m$. Figure \ref{fig:mainfig}.(IV) shows the decoders, $Q^{image}$ and $Q^{joint}$, and predicted distributions, $(\mu_{t_{target}}^{image},\sigma_{t_{target}}^{image})$ and $(\mu_{t_{target}}^{joint},\sigma_{t_{target}}^{joint})$, for two domains. The learning objective of our framework is to construct better distributions according to the given observations as in \cite{CNP} and \cite{sekerconditional}, so the loss term is defined as:
\begin{align}
 \mathcal{L} = - \sum_m^M logP(S^m_{t_{target}}\mid \mu^m_{t_{target}},\sigma^m_{t_{target}})
\end{align}
where $S^m_{t_{target}} \in S^M_{t_{target}}$ is the target sensorimotor value for modality $m$ at time $t_{target}$.

After training, the system can be requested to make predictions for all the modalities and for all the time-steps by fixing \erhanadd{$p^M=1/M$} and assigning $O^M$ as novel observations. By observing the sensorimotor state at any time-step, any other time point before and after can be queried and predicted using our framework. According to the situation, if a sensorimotor modality does not seem to provide reliable signals, the weight given to that modality can be decreased by configuring availability vector $w$. Note that, the system can even predict missing modalities if the corresponding $w^m$ is set to zero because of a lack of the modality.  Our framework can use the shared latent space for multi-modal predictions. This also enables our framework to imitate other agents by observing their actions with, for example, vision and sound, and producing the agent own behaviour by predicting the corresponding motor signals.

\section{Experiment Setup}
\label{sec:expsetup}

\begin{figure}[h]
    \centering
    \includegraphics[width=\linewidth]{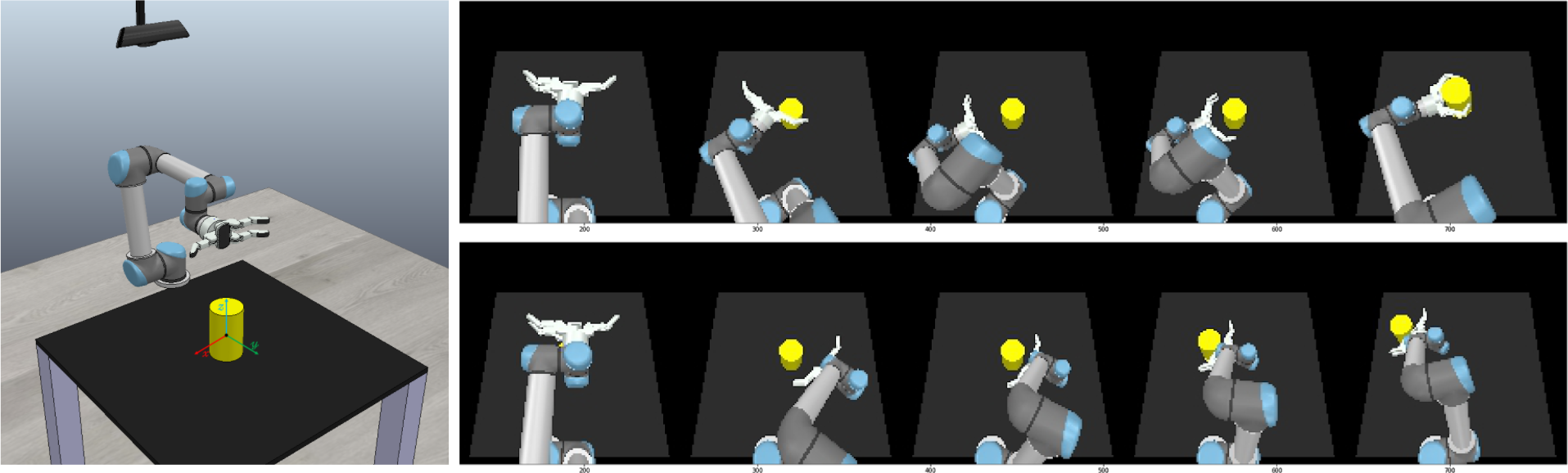}
    \caption{(Left) Experiment setup with vision sensor , UR5, and the object at the middle of the table. (Right) Example grasping and pushing actions recorded via the vision sensor.}
    \label{fig:expsetup}
\end{figure}

To demonstrate the capabilities of our system, we designed an experiment where the actions of the robot can be predicted from the visual and proprioceptual observations at the beginning of the movement execution. A simulated environment was built using CoppeliaSim \cite{coppeliaSim}. The setup consisted of a UR5 robot equipped with a three-finger gripper, a vision sensor, and an object on a table to be manipulated by the robot (Fig. \ref{fig:expsetup} left). The action repertoire of the robot was composed of parameterized push and grasp actions that allow reaching to the object from all directions, and the data collection protocol for each action execution (interaction) was as follows. At the beginning of each interaction, the robot initialized its wide-open hand at an initial position, and an object appeared in the middle of the table (Fig. \ref{fig:expsetup} right). If the selected action was push, a random pushing angle was sampled and the robot pushed the object from this angle to a predetermined fixed distance of 30cm while keeping the hand open. If the selected action was grasp, a random grasping angle was sampled and the robot started to close its hand while approaching to the object so as to grasp it and lift it to a fixed height over the table (30cm). The collected data consisted of two modalities that are $proprioception$ and $vision$. The proprioceptive signals were composed of seven joint angles of the robot (6 joints of the UR5 robot and 1 hand opening joint), whereas the visual signals were 128 x 128 x 3 RGB images. Visual signals \erhan{were} collected via the vision sensor that \erhan{was} placed to the point of view of the robot (see Figure \ref{fig:expsetup}). In the end, 50 successful push and grasp interactions (100 in total) were collected using the simulator. The interactions \erhan{were}  separated into train and test set with 80\% and 20\% ratios respectively.
 
\section{Experimental Results}
\label{sec:expresult}

We conducted a set of experiments to test the capabilities of \methodname~from different aspects. First, in Section \ref{subsec:pred}, we verify the prediction capabilities of \methodname~by generating complete image and joint trajectories conditioned only on single images. In Section \ref{subsec:missing}, the performance of \methodname~ is compared with MVAE and multi-step errors made by these models are analyzed in Section \ref{subsec:longhorizon}.
%As \methodname~can make independent predictions for future timesteps, it has a lower error compared to MVAE; the error does not cascade. 
%We analyze the multi-step errors of \methodname~and MVAE to show this difference i 
In Section \ref{subsec:tsne},  we show how the latent space of two modalities indeed blend with each other. In Section \ref{subsec:mirror}, we analyze the behavior of our model when conditioned with images from different perspective and whether it can serve as a mirror neuron system  in replicating observations from different agents. We analyze whether such generalization is due to the inductive bias of the model by making two different ablation studies in Sections \ref{subsec:knn} and \ref{subsec:onlyimage}, together which lend support to the the idea that mirror neuron formation can be mediated by self-observation and modality blending with \methodname. Lastly, we test the generalization of the model by conditioning on out-of-distribution samples and include the results in \ref{subsec:generalization}.

\subsection{\erhanadd{Long-term Prediction with Vision only}}
\label{subsec:pred}

\begin{figure*}[t]
    \centering
    \includegraphics[width=\linewidth]{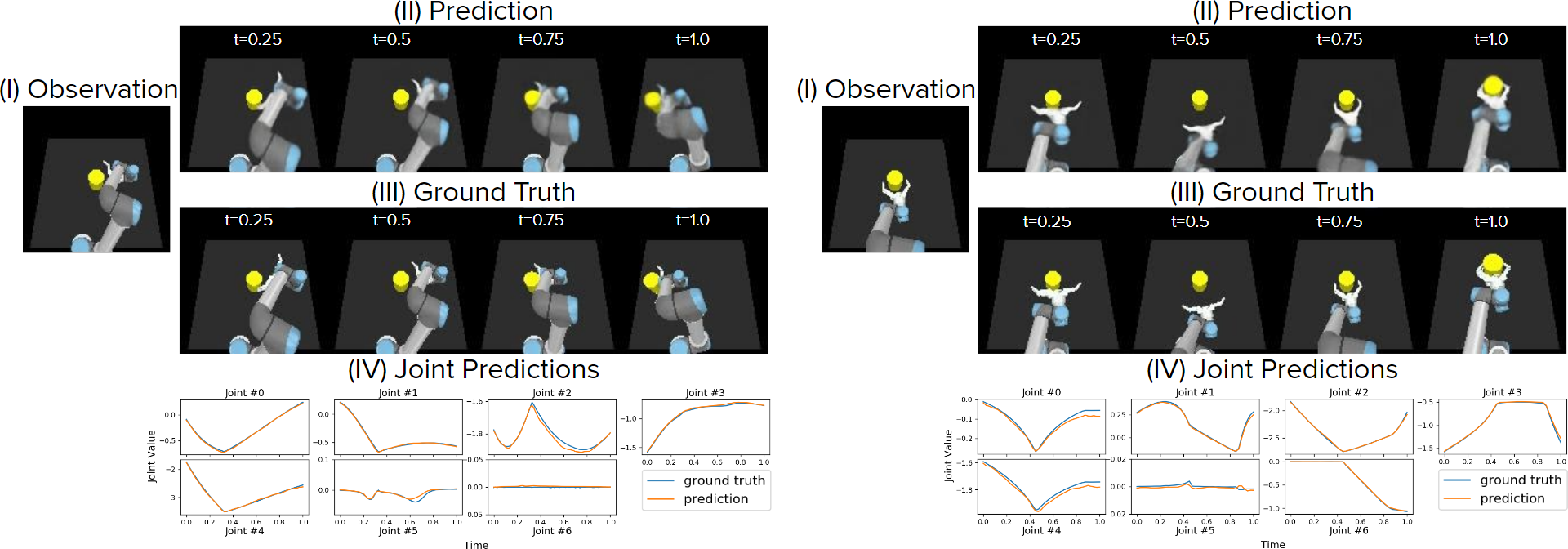}
    \caption{(I) Images that are used as observations. (II) \methodname~visual predictions for the given time-steps. (III) Ground truth images for the given time-steps. (IV) \methodname~7D joint predictions for the whole action}
    \label{fig:expresult}
\end{figure*}

In this experiment, we verify whether our system can produce visual sequences and the corresponding joint values given a single image as an input. Note that since we take an average of latent vectors for conditioned points, we might as well give multiple images, instead of a single one, to get a more accurate prediction (see Equation \ref{eq:avg_latents}). Here, to demonstrate the capabilities of our system even in such a scenario where the information is minimum, the system is fed with a single visual observation which is obtained just before the robot interacts with the object. The availability vector is set to one for visual modality and to zero for proprioceptive modality since the observation only includes visual information. Then, the system is requested to produce visual and motor signals from the beginning to the end of the movement.

% \erhcom{I think it is better to explain -again- that any number of samples can be used to condition the process, and then mention that to relate the results with manipulation and mirroring we use a singe frame before contact}

Figure \ref{fig:expresult} shows two examples of pushing and grasping actions at the left and the right of the figure respectively. Figure \ref{fig:expresult}.(I) shows the obtained images that are used as observations from the test set. Figure \ref{fig:expresult}.(II-III) shows the predicted images together with the ground truth at the corresponding time-steps. It can be seen that exploiting the position, orientation, and hand state of the robot extracted from the observed image, our system could successfully predict the sequences of \erhann{visual and  proprioceptive signals} from start to finish, which are highly accurate compared to the ground truth \erhann{values}. 
%Figure \ref{fig:expresult}.(IV) also shows the prediction results for the proprioceptive modality. 
\erhann{It is notable that d}espite the fact that there was no proprioceptive observation in these two examples, our model could make accurate joint predictions from start to the end of the movement by just \erhann{having access to visual modality (see Figure \ref{fig:expresult}.(IV))}. These results indicate that our model can use the representation encoded from an available modality to predict the signals of the other missing modalities. %Imitation performance based on the predicted joint values are presented in the. 
A more detailed quantitative analysis about cross-modality predictions \erhann{is presented} in the next section.

%\begin{figure}[h]
%    \centering
%    \includegraphics[width=\linewidth]{figures/joint.png}
%    \caption{Example cross-modality joint predictions of our framework by using a single image as the observation.}
%    \label{fig:jointprediction}
%\end{figure}

\subsection{\erhann{Missing Modality Prediction  as a Function of Training Set Size}}
\label{subsec:missing}

\begin{figure}[h]
    \centering
    \includegraphics[width=\linewidth]{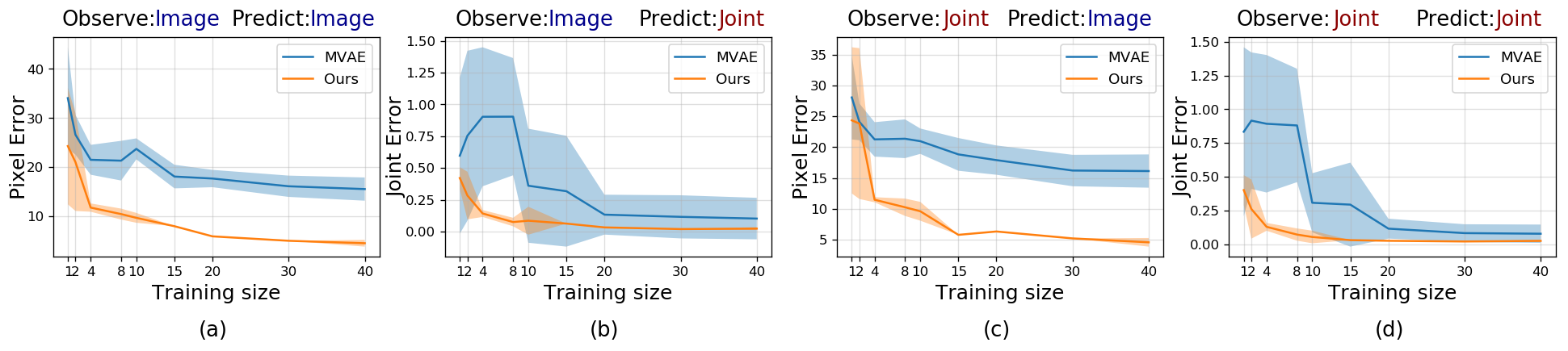}
    \caption{The prediction errors on the test set for different modality input-output pairs with the increasing size of the training data (x-axis).}
    \label{fig:partialobservation}
\end{figure}

In this section, we test whether \methodname~can indeed perform well when there are some missing modalities. In this experiment, we used the same network in the previous experiment which is trained \erhann{ by using either} $visual$ \erhann{or} $proprioceptive$ modalities. During the test phase, we set the availability of one of the two modalities to zero. We tested whether our system can still predict missing modalities.

We compared our method with MVAE \cite{zambelli2020multimodal} as it can handle missing modalities. We made several modifications to the original MVAE architecture to make a fair comparison. First, we added convolutional layers for the visual input pipeline. All layers in the encoder and the decoder are exactly the same as in \methodname. Therefore, the number of parameters is the same except that MVAE uses an extra fully-connected layer to combine different encoder outputs. This extra layer is not needed in \methodname~since the latent representation is shared and acquired via normalized weighted summation. Second, we remove the standard deviation prediction from the decoder as it gave better results in our preliminary experiments. We did not use the KL divergence term in the loss as in \cite{zambelli2020multimodal}. Third, we randomly mask the sensorimotor data at time $t$ and predict the data at $t+1$, in addition to other masking schemes reported in \cite{zambelli2020multimodal}. This additional masking scheme enables us to make full trajectory predictions (both forward and backward prediction) given the observation before contact. Our implementation\footnote{\url{https://github.com/alper111/multimodal-vae}} is based on \cite{zambelli2020multimodal} and their code repository\footnote{\url{https://github.com/ImperialCollegeLondon/Zambelli2019_RAS_multimodal_VAE}}.

We report our results in Figure \ref{fig:partialobservation} where the prediction accuracies with increasing number of training trajectories are shown. For the two modalities in our experimental setup, we tested four different combinations of modality masking: predicting visual states when either proprioceptive modality (Figure \ref{fig:partialobservation}.a) or the visual modality (Figure \ref{fig:partialobservation}.c) is missing, and predicting joint states when either proprioceptive modality (Figure \ref{fig:partialobservation}.b) or the visual modality (Figure \ref{fig:partialobservation}.d) is missing. We condition both \methodname~and MVAE models with the observations taken from the same time-step that is right before the robot interacts with the object. Both systems predict complete visual and joint trajectories starting from $t=0$ to $t=T$. Since \methodname~is able to learn from few data, the error and its variation drop quickly even with the small training size, and it improves the accuracy while the data size is increased. For MVAE, the error slightly drops during the data size increase, yet, still far from \methodname. One reason for the error of MVAE is that it feeds the predictions back to itself as input, thus cascades the error in the long horizon. We investigate this phenomenon in the next section in detail.
% erhan: I am confused, doesnot MVAE need a sequence of input to properly function?
% alper: Given data at t, MVAE can predict t+1 and t-1 with our training scheme. By feeding the prediction as input, we can further predict t+2 and t-2 and so on.

\subsection{Analysis of Long Horizon Predictions}
\label{subsec:longhorizon}

\begin{figure}[h]
    \centering
    \includegraphics[width=\linewidth]{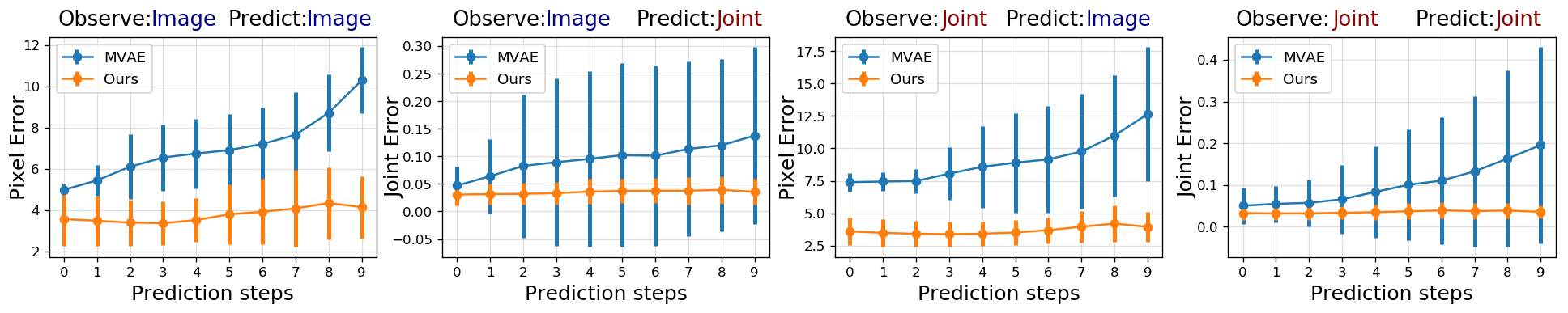}
    \caption{Multi-step prediction results. MVAE errors increase with the prediction steps due to error accumulation. However, our model preserves the error at the same level for increasing prediction steps since it predicts every timestep independent from each other}
    \label{fig:multistep}
\end{figure}

In this section, we compared the capacity of \methodname~on the long horizon predictions with the MVAE method. Both models are trained using the same two modalities in the same way as in the previous section.

In \cite{zambelli2020multimodal}, MVAE is used for one step ahead predictions to control the iCub humanoid robot in a closed loop. To make predictions about further time-steps, the model can be fed with its output from the previous time-step. They showed that when trained with sinusoidal data, the prediction accuracy remains the same for about 50 time-steps, and then starts to degrade. In this experiment, we compared the two methods using the data that is collected during the self-exploration which is more complex and high-dimensional. In contrast to MVAE, \methodname~does not need to feed its output back to itself as input to make further predictions since we can explicitly query any time-step independently and make predictions on the long horizon directly.

We analyze the error versus the prediction step for two methods in Figure \ref{fig:multistep}. The error of MVAE increases as the prediction step increases since the error is fed back in the input for future time-step predictions. However, the error of \methodname~remains around the same because the model does not have a feedback loop to connect an erroneously computed output to its input, and make predictions for every time-step independently just by looking to the observations.

\subsection{Multimodal latent space visualization}
\label{subsec:tsne}
% \erhcom{This is too low level result to be the first of the Experimental results, let's move it to just before Mirror section}

 \begin{figure}[h]
    \centering
    \includegraphics[width=\linewidth]{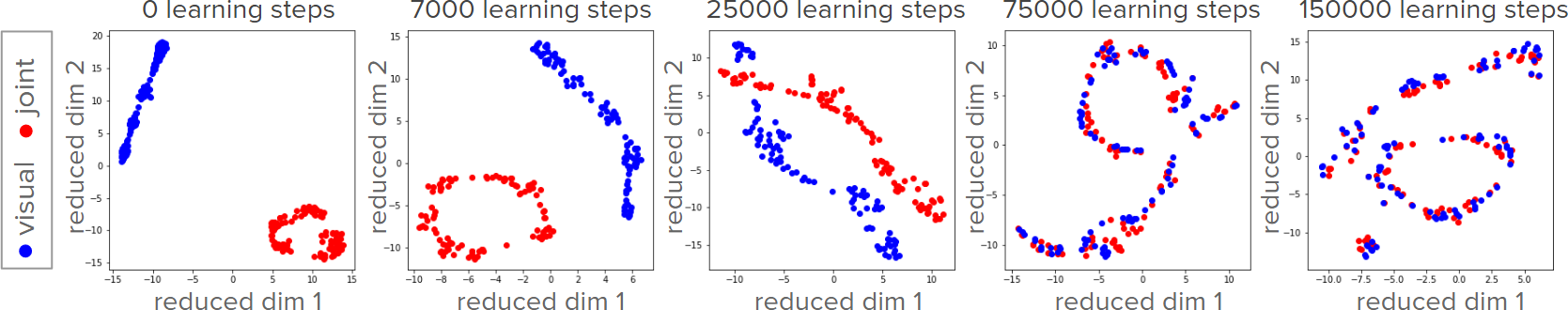}
    \caption{t-SNE visualization of latent space during training. Blue points are visual encodings, and red points are joint encodings.}
    \label{fig:representation}
\end{figure}

In this experiment, multi-modal latent space is visualized and analyzed. As mentioned in the previous section, we trained\footnote{Training details about the network can be found in \ref{apx:DMBN}.} the system with the two modalities that are $visual$ and $proprioceptive$. For visualization purposes, the high-dimensional representation space (128 sized vector) is reduced to two dimensions using t-SNE \cite{tsne} method at different stages of the training. Figure \ref{fig:representation} shows the t-SNE visualization of the multimodal latent space at 0, 7k, 25k, 75k, and 150k learning steps from left to right. Blue and red points indicate the samples from the visual and proprioceptive modalities, respectively. Figure \ref{fig:representation} shows that although the different modalities are clustered and separated from each other at the beginning of the training (0 and 7k learning steps), they start to share the representations between each other after a while (25k learning steps), and turn into matching/overlapping representations in the later stages of the training (75k and 150k learning steps). Paired blue and red points in the overlapping representation space are analyzed and it is found that each paired blue-red point corresponds to two modalities recorded from the same state of the environment. These results suggest that our system can effectively learn multiple modalities in a common latent space in a way that every sensorimotor modality recorded from the same state of the environment ends up turning into the nearly same representation in the latent space. This allows our system to predict the missing modalities by using the representations produced by other available modalities, which was shown in the Section \ref{subsec:missing}.

\subsection{Imagining Own Actions by Observing Others: Emergence of Mirror Neuron System Behaviour}
\label{subsec:mirror}

\begin{figure}[h]
    \centering
    \includegraphics[width=\linewidth]{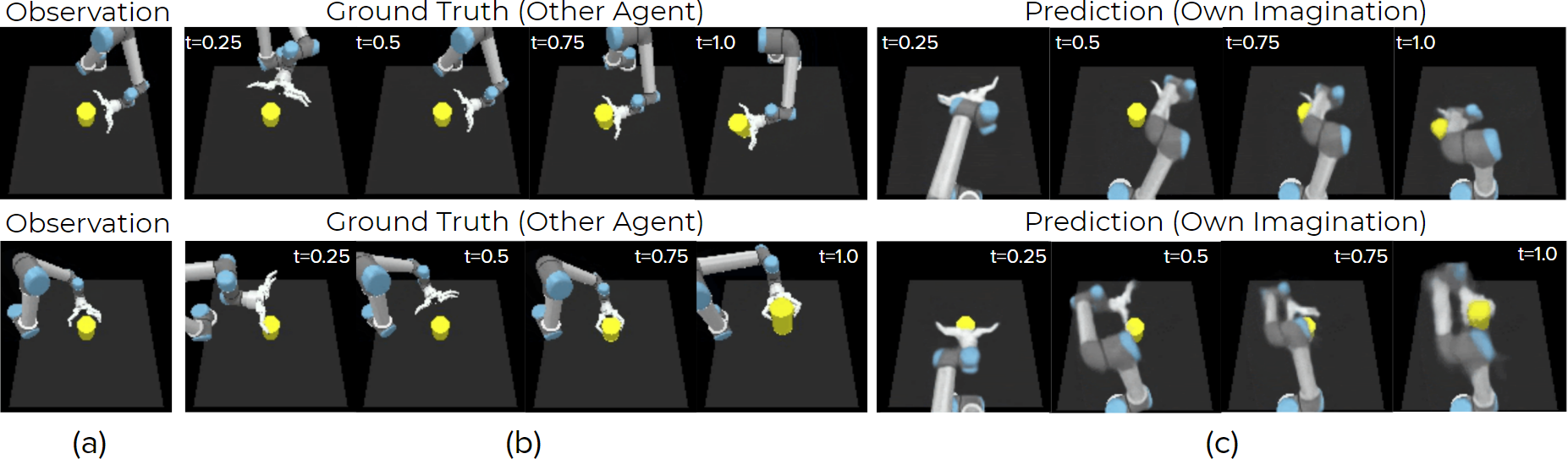}
    \caption{Examples of \methodname~effect imitation behavior. First row: Observing the other agent just before it pushes away the object. Second row: Observing the other agent just before it grasps the object.}
    \label{fig:effectimagine}
\end{figure}

\begin{figure}[t]
    \centering
    \includegraphics[width=\linewidth]{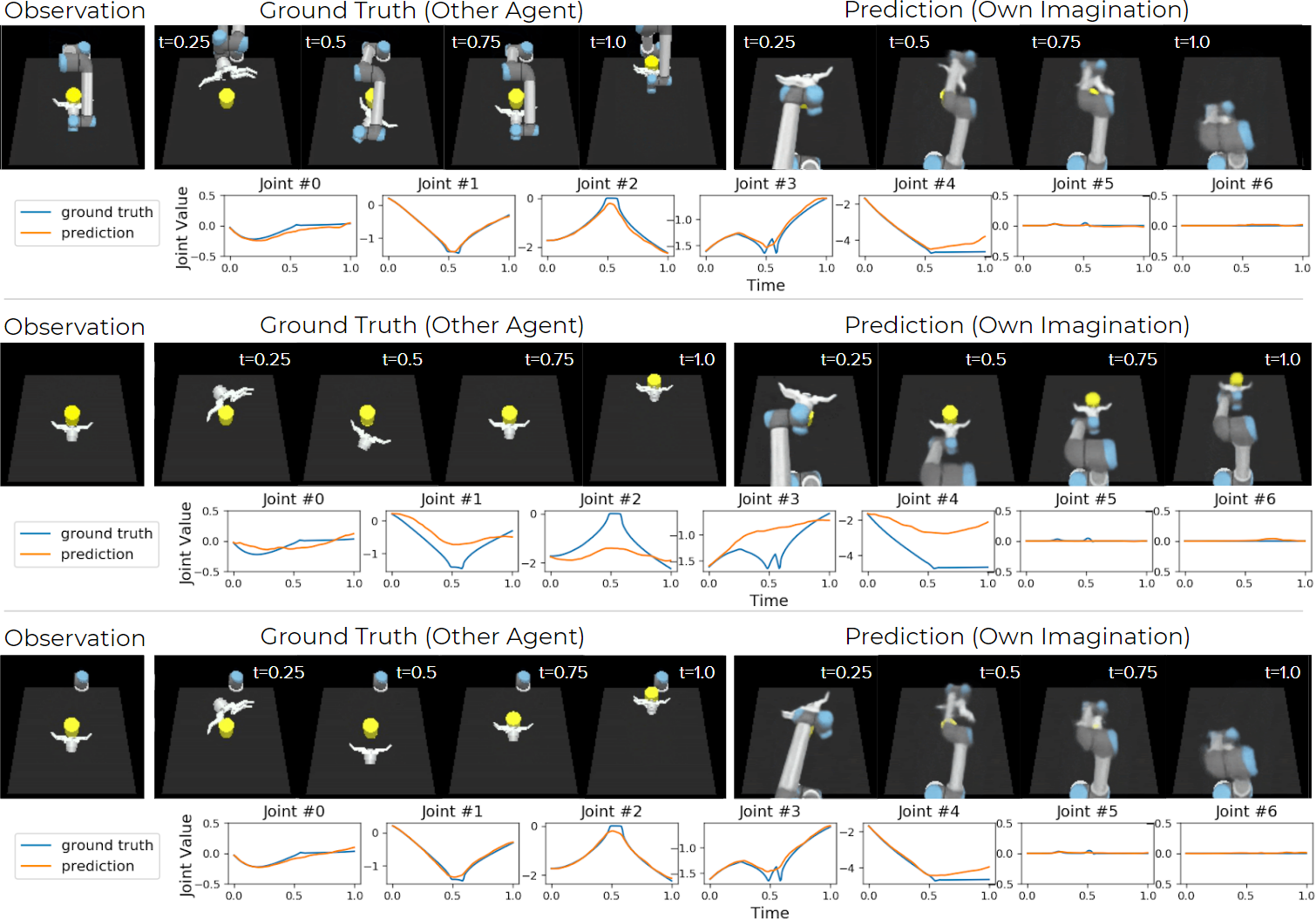}
    \caption{Examples of \methodname~egocentric imitation behavior. First row: Emergence of mirror neuron behaviour where the agent observes the other agent pull the object towards itself. Second row: The agent observes a hand without the body. Third row: The agent observes a hand and a base without the arm.}
    \label{fig:mirrorimagine}
\end{figure}

In this experiment, we tested our system to see if it can generate own sensorimotor data by observing another agent perform an action. In order to do that, an agent was placed on the different sides of the table and their performed actions are observed via our agent's visual sensor. Note that in the training data, interactions were only performed and recorded just by our agent, so observing other agents in the test time is a novel information which is completely outside of our training set. Since we were using only visual data as the observation, availability vector is set to one for visual modality and to zero for proprioceptive modality. Because of the fact that the observations are on another agent but the predictions are made for our agent, this prediction process can be considered as the imagination \erhan{of the action of another agent for the self}.

Figure \ref{fig:effectimagine} shows the prediction results of our model in two different pushing and grasping scenarios where the observations are shown in Figure \ref{fig:effectimagine}(a). In the first scenario, the other agent was placed on the opposite side of the table, and in the second scenario, the other agent was placed on the left side of the table. Figure \ref{fig:effectimagine}(b) shows the visual signals during the other agent performed its action, and Figure \ref{fig:effectimagine}(c) shows the full trajectory prediction of our system as it imagines the visual signals for itself. As it can be seen in the predictions, our agent is able to generate visual trajectories from its own perspective that matches the approaching angle and the action type in the observation, hence, \erhan{imagining an action that would be an effect-based imitation of the observed action} %\erhcom{there is no actual execution so we cannot say there is imitation. I fixed this section with imagining an action that would...}
% \erhcom{Onemli: burada joint angle predictionlarini actor joint angle'lari ile kesin karsilastirmamiz gerek; Figure 4 ile ayni sekilde yani}. \yunus{(Effect prediction yaptiklari icin farkli olmayacaklar mi ?)}

However, when we further analyzed our model, we saw that \methodname~behaves differently in some specific scenarios. Surprisingly, when the other agent pulls the object towards itself , our agent \erhan{imagines an action} that egocentrically imitates the \erhan{observed} action (Figure~\ref{fig:mirrorimagine}, first row) and generates motor signals that \erhan{would} also pull the object towards itself \erhan{rather than creating the effect on the object} as shown in the Figure \ref{fig:effectimagine}. 
\erhan{We can say that, in this particular action observation case, an emergent mirror neuron property was exhibited by our DMBN. Interestingly this behavior `switches' so that the action imagined corresponds to effect-based imitation (i.e. \textit{emulation}) of the observed action when the effector is removed from the interaction (Figure~\ref{fig:mirrorimagine}, second row). Finally, when the robot is partially revealed by disclosing the base of the robot, the system starts to understand the observed action again as bringing the object toward one's self (Figure \ref{fig:mirrorimagine}, third row), thereby showing a mirror neuron response as in the first row of Figure~\ref{fig:mirrorimagine}}.
% \erhcom{joint angles should be shown}.

These results show that when conditioned with the visual signals of other agents, \methodname~  \erhan{has potential to produce output signals similar to that of a mirror neuron system. However,  signals generated can correspond to either effect-based or egocentric imitation depending on the specific visual signals available from the other agent and the environment. Therefore, it is viable to modulate the behavior of DMBNs via other cognitive mechanisms, e.g. attention, to purposefully control the operation of the model.}
% \erhcom{The effect-based imitation can sometimes happen with full vision also, right? Do not we mention about this??}\yunus{(didn't Figure 8 show that or ?)}

\subsection{Template Matching According to Pixel and Latent Space Distances}
\label{subsec:knn}

\begin{figure}[h]
    \centering
    \includegraphics[width=\linewidth]{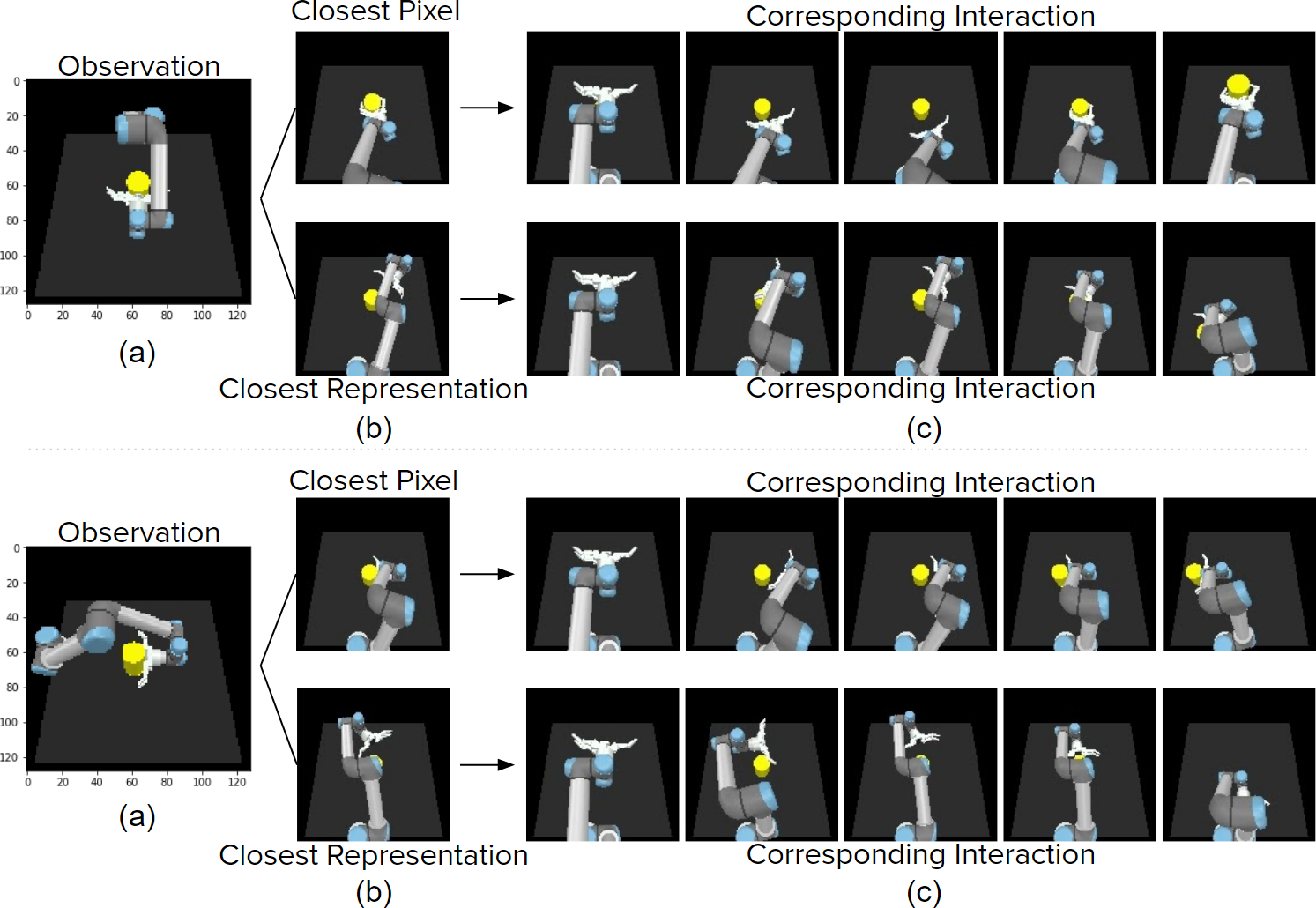}
    \caption{Closest Pixel: Pixel Space Distance;   Closes Representation: Latent Space Distance }
    \label{fig:temp_matching}
\end{figure}

In this experiment, \erhan{we aimed to see whether the mirror neuron emergence in our system was due to the rich representations constructed in the latent space during the learning, or it could be simply explained by a straightforward image based template matching. For this, first, two test cases in which true mirror response (i.e. action representation that would yield egocentric imitation) was observed was selected.} Figure \ref{fig:temp_matching}.(a) shows these two observation cases where the agents are placed on the opposite and the left side of the table, respectively. For each test case, the closest image in the training set that gives the minimum average pixel error, and the corresponding image of the closest representation that gives the minimum MSE error in the representation latent space are found and compared. Figure \ref{fig:temp_matching}.(b) shows the corresponding closest pixel and representation images for each case respectively. Figure \ref{fig:temp_matching}.(c) also shows the corresponding full trajectory interactions of the found results from the training set.

Results of the both examples show that the corresponding interactions of the closest pixel images do not \erhan{exhibit true mirror response (i.e. the predicted signals would not yield an egocentric imitation when executed on the robot)}. On the other hand, the corresponding interactions of the closest \erhan{latent space} representations show \erhan{true mirror response}. These results suggest that the output signals that \methodname~produces are not based on a simple image based error minimization but on rich representations that are learned during the multi-modal training \erhan{with modality blending}. \erhan{The} contribution of the deep modality blending to the mirror neuron emergence is further inspected in detail in the next section.

\subsection{Analysis of the Contribution of Multimodal Learning to Mirror Neuron Emergence}
\label{subsec:onlyimage}

\begin{table}[h]
\centering
\resizebox{0.55\linewidth}{!}{%
\begin{tabular}{|c|c|c|c|}
\hline
\multicolumn{2}{|c|}{\erhan{True Mirror Response}} &
  Success &
  Fail \\ \hline
 &
  Case 1 &
  10 &
  0 \\ \cline{2-4} 
\multirow{-2}{*}{\begin{tabular}[c]{@{}c@{}}Image + Joint Model\end{tabular}} &
  Case 2 &
  10 &
  0 \\ \hline
 &
  Case 1 &
  6 &
  4 \\ \cline{2-4} 
\multirow{-2}{*}{\begin{tabular}[c]{@{}c@{}}Only Image Model\end{tabular}} &
  Case 2 &
  4 &
  6 \\ \hline
\end{tabular}%
}
\caption{Test results of the two models in ten different training \erhan{sessions} with two action observation cases (see Figure~\ref{fig:temp_matching}) of demonstrating agent positioned across (Case 1) and the left side of the agent (Case 2) . Success: The model produced a signal output that corresponds to true mirror response; i.e. the execution of the action based on those signals would yield egocentric imitation. Fail: the model produced disturbed image signals}
\label{tab:my-table}
\end{table}

In this experiment, we tested if deep modality blending contributes to the mirror neuron emergence in our system. To do that, a model that only uses visual modality was trained next to our model which was trained by using both visual and proprioceptive modalities. In order to prevent the training biases that can occur because of the initial network weights or sampling seeds, both models were trained 10 times with different different random initializations. After the training, both of the models were tested with two test cases and checked whether the networks produce output signals that correspond to mirror neuron emergence. \erhan{The two test cases used in this experiment were the same examples as in the Experiment \ref{subsec:knn} where the demonstrating agent were placed at the opposite and the left side of the table (see Figure~\ref{fig:temp_matching}).}

\begin{figure}[h]
    \centering
    \includegraphics[width=0.8\linewidth]{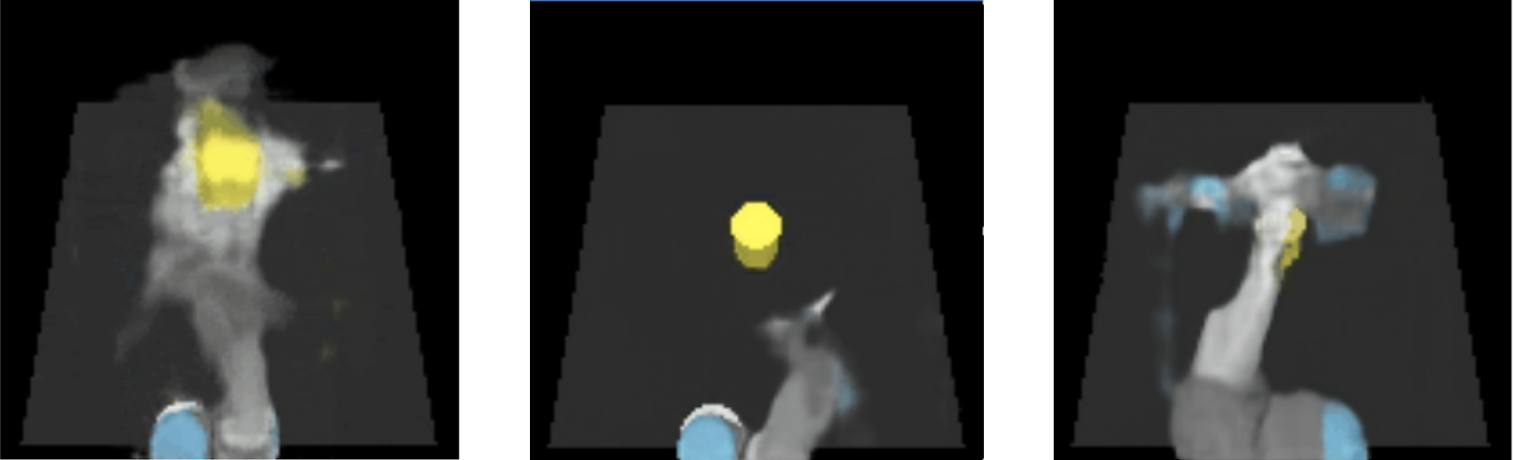}
    \caption{Example failing scenarios for the only image model. The images are disturbed and the robot arm is disappearing. 
    %\erhcom{Only Image Model de output yine Image+Joint degil mi?}
    }
    \label{fig:failcases}
\end{figure}

Table \ref{tab:my-table} shows the results of the two models in ten different training initializations with two test cases. Results indicate that the model that uses deep modality blending (the model with Image + Joint) produces coherent images that corresponds to egocentric imitation in every test case where the model that uses only one modality (Only Image Model) produces disturbed images 
% \erhcom{EDIT. here you are mixing image output and imitation signal=motor output..}
on the ten test cases out of twenty. Figure \ref{fig:failcases} shows some example fail cases for the only image model where the image is disturbed or the arm of the robot is disappeared 
%\erhcom{EDIT. Please let's see the joint predictions as well!}. \yunus hocam bu sadece image modality ondan gosteremiyoruz.
These results suggest that using deep modality blending with visual and proprioceptive modalities contribute to the emergence of mirror neuron behavior.

%\section{Discussion}
%\label{sec:discussion}
%+ modality blending burda key nokta olmalı, çünkü farklı modalityler bir şekilde aynı şeye refer %edebilir durumda olmalı. çünkü mesela insan kör olunca birden tüm cognitive skillerini kaybetmiyor.
%+ burada vision+jointi blendliyoruz, ama aslında bu birebir bir fonksiyon gibi. Mesela ses modalitysi %olsaydı, piyano tuşuna bassaydık, tuşa basana kadar ses 0, basınca ses 1. Ses 0 conditionı verip joint sonucu ne gelecekti? buradan ilginç sonuçlar çıkabilir mi? çünkü ses discrete bir representation olcak. aynı şekilde haptic modality.
%- robottan bağımsız ve hareketli bir şey yok.
%Mirror imitation vs Egocentric Imitation, conditions that specifies which behavior will emerge.

\section{Conclusion}
\label{sec:conc}

\erhan{In this work, we proposed Deep Multi-modal Blending Network (DMBN) as a multi-modal action representation system that learns the sensorimotor signals corresponding to the actions, in a robust latent representation allowing temporal cross-modal predictions with limited information. DMBNs can generate complete  signal trajectories in any desired  modality even with zero information on the desired modality by using other available modalities. The performance of the network surpasses the available multi-modal learning systems due to long-range one-shot prediction capability and its novel stochastic modality blending mechanism.}

\erhan{DMBNs  build powerful internal representations that allow surprisingly dynamic extrapolation properties, making it a strong contender as a feature-engineering-free Mirror Neuron System model. To be specif, after learning proprioception and visual signals based on self action observations, when tested with different perspective action observations, it successfully generates valid signals that represents its own actions. Depending on the visual setting, the network either acts a true mirror system matching an observed act to its own repertoire in an egocentric way, or acts as an effect-based action matching system. Thus, the network has potential to sustain egocentric  and effect-based action recognition and imitation capabilities when envisioned in the cognitive system of an artificial or biological agent.}

\erhan{In this vein, future work should focus on developing biologically plausible and developmentally realistic end-to-end mirror neuron systems that learn along with sensorimotor skill acquisition. In the current study, we used a fixed action repertoire to systematically study the properties of DMBNs; yet in a developing artificial or biological cognitive agent, mirror neuron formation and action learning should go in parallel creating potentially non-trivial interactions worth studying. Another direction that should be pursued is to use the basic imitation capacity acquired by the model, to construct \textit{novel imitation} capacity, where the parts of an observed novel act can be understood in terms of and matched to the existing action repertoire of the agent with the help of DMBN implementing the developing mirror neuron system. We believe that work around these directions will not only stimulate computational  study of mirror neurons as a full end-to-end system but also form a framework for lifelong sensorimotor learning for social robots.}

\section*{Acknowledgement}
This  research  has  received  funding  from  the  European Union's  Horizon  2020  research  and  innovation  programme under grant agreement no. 731761, IMAGINE; was   partially   supported   by   JST   CREST ``Cognitive   Mirroring'' under grant no. JPMJCR16E2, by the International Joint Research Promotion Program of Osaka University  under the project ''Developmentally and biologically realistic modeling of perspective invariant action understanding'' and by the Turkish Directorate of Strategy and Budget under the TAM Project number 2007K12-873. The numerical calculations reported in this paper were partially performed at TUBITAK ULAKBIM, High Performance and Grid Computing Center (TRUBA resources).

\bibliography{mybibfile}

\appendix

\section{Generalization of the System to the Novel Environmental Configurations}
\label{subsec:generalization}

\begin{figure}[h]
    \centering
    \includegraphics[width=\linewidth]{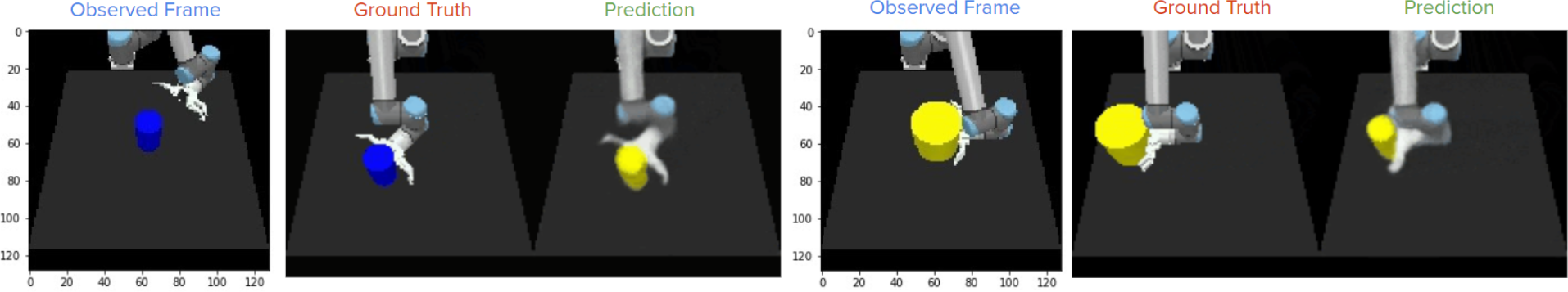}
    \caption{Generalization performance of the proposed system in two different configurations. Left side: the color of the object is blue. Right side: the size of the object is bigger than the original one.}
    \label{fig:expgeneral}
\end{figure}

In this experiment, we tested our system with different novel environmental configurations that have different properties than the training set. Figure \ref{fig:expgeneral} shows the generalization performances of the two different configurations. Left side of the figure shows a scenario in which the color of the object was different from the object in the training data, and the right side shows a configuration where the size of the object was changed. Despite not seeing a big or blue object in the training, our system could successfully predict the correct approaching angle and the action using the observed image in both configurations. It can be seen that the color and the size of the objects are predicted as in the configuration in the training images. This is expected since the only configuration for object in the training scene was yellow and small. Even though the object in the observed image was not the same with the training object, our system could use the knowledge that is learned in the training data to predict a correct output in its own configurations that satisfies the given observation.

\section{Network Architecture and Training Details of DMBN}
\label{apx:DMBN}

In this section, the network architecture and training configurations of DMBN are shown. Table \ref{tab:DMBNImageEncoder} and Table \ref{tab:DMBNJointEncoder} show the image and joint encoder architectures respectively. Table \ref{tab:DMBNImageDecoder} and Table \ref{tab:DMBNJointDecoder} show the image and joint decoder architectures respectively. DMBN is trained with Adam optimizer \cite{kingma2014adam} for one million iterations with a batch size of one and a learning rate of 0.0001. We set $obs_{max}$ to 5.

\begin{table}[h]
    \centering
    \resizebox{\linewidth}{!}{
    \begin{tabular}{|c|c|c|}
        \hline
        Layer & Input size & Output size \\
        \hline
        \hline
        Conv3x3 + ReLU + MaxPool2x2 & (4, 128, 128) & (32, 64, 64)\\
        \hline
        Conv3x3 + ReLU + MaxPool2x2 & (32, 64, 64) & (64, 32, 32)\\
        \hline
        Conv3x3 + ReLU + MaxPool2x2 & (64, 32, 32) & (64, 16, 16)\\
        \hline
        Conv3x3 + ReLU + MaxPool2x2 & (64, 16, 16) & (128, 8, 8)\\
        \hline
        Conv3x3 + ReLU + MaxPool2x2 & (128, 8, 8) & (128, 4, 4)\\
        \hline
        Conv3x3 + ReLU + MaxPool2x2 & (128, 4, 4) & (256, 2, 2)\\
        \hline
        Flatten & (256,2,2) & 1024\\
        \hline
        Dense & 1024 & 128 \\
        \hline
        Multiply (Image Coefficient) & 128 * 128 & 128 \\
        \hline
    \end{tabular}
    }
    \caption{\methodname~Image Encoder}
    \label{tab:DMBNImageEncoder}
    \vspace{-0.25cm}
\end{table}

\begin{table}[h]
    \centering
    \resizebox{0.8\linewidth}{!}{
    \begin{tabular}{|c|c|c|}
        \hline
        Layer & Input size & Output size\\
        \hline
        \hline
        Dense + ReLU & 8 & 32\\
        \hline
        Dense + ReLU & 32 & 64\\
        \hline
        Dense + ReLU & 64 & 64\\
        \hline
        Dense + ReLU & 64 & 128\\
        \hline
        Dense + ReLU & 128 & 128\\
        \hline
        Dense + ReLU & 128 & 256\\
        \hline
        Dense + ReLU & 256 & 128 \\
        \hline
        Multiply (Joint Coefficient) & 128 * 128 & 128 \\
        \hline
    \end{tabular}
    }
    \caption{\methodname~Joint Encoder}
    \label{tab:DMBNJointEncoder}
\end{table}

\begin{table}[h]
    \centering
    \resizebox{\linewidth}{!}{
    \begin{tabular}{|c|c|c|}
        \hline
        Layer & Input size & Output size\\
        \hline
        \hline
        Add (Image + Joint Representations) & 128 + 128 & 128\\
        \hline
        Concatenate (Target Layer) & 128 & 129\\
        \hline
        Dense + ReLU & 129 & 1024\\
        \hline
        Reshape & 1024 & (256, 2, 2)\\
        \hline
        Conv3x3 + ReLU + UpSample2x2& (256, 2, 2) & (256, 4, 4)\\
        \hline
        Conv3x3 + ReLU + UpSample2x2& (256, 4, 4) & (128, 8, 8)\\
        \hline
        Conv3x3 + ReLU + UpSample2x2& (128, 8, 8) & (128, 16, 16)\\
        \hline
        Conv3x3 + ReLU + UpSample2x2& (128, 16, 16) & (64, 32, 32)\\
        \hline
        Conv3x3 + ReLU + UpSample2x2& (64, 32, 32) & (64, 64, 64)\\
        \hline
        Conv3x3 + ReLU + UpSample2x2& (64, 64, 64) & (32, 128, 128)\\
        \hline
        Conv3x3 + ReLU & (32, 128, 128) & (16, 128, 128)\\
        \hline
        Conv3x3 + ReLU & (16, 128, 128) & (8, 128, 128)\\
        \hline
        Conv3x3 + Sigmoid & (8, 128, 128) & (3, 128, 128)\\
        \hline
    \end{tabular}
    }
    \caption{\methodname~Image Decoder}
    \label{tab:DMBNImageDecoder}
\end{table}

\begin{table}[h]
    \centering
    \resizebox{\linewidth}{!}{
    \begin{tabular}{|c|c|c|}
        \hline
        Layer & Input size & Output size\\
        \hline
        \hline
        Add (Image + Joint Representations) & 128 + 128 & 128\\
        \hline
        Concatenate (Target Layer) & 128 & 129\\
        \hline
        Dense + ReLU & 129 & 1024\\
        \hline
        Dense + ReLU & 1024 & 512\\
        \hline
        Dense + ReLU & 512 & 216\\
        \hline
        Dense + ReLU & 216 & 128\\
        \hline
        Dense + ReLU & 128 & 32\\
        \hline
        Dense & 32 & 14\\
        \hline
    \end{tabular}
    }
    \caption{\methodname~Joint Decoder}
    \label{tab:DMBNJointDecoder}
\end{table}

\section{Network Architecture and Training Details of MVAE}
\label{apx:MVAE}

In this section, the network architecture and training configurations of MVAE are shown. Table \ref{tab:MVAEImageEncoder} and \ref{tab:MVAEJointEncoder} show the image and the joint encoder architectures respectively. Table \ref{tab:MVAEShared} shows the shared encoder-decoder architecture. Table \ref{tab:MVAEImageDecoder} and \ref{tab:MVAEJointDecoder} show the image and joint decoder architectures respectively. MVAE is trained with Adam optimizer \cite{kingma2014adam} for 200 epochs with a batch size of 128 and a learning rate of 0.001.

\begin{table}[h]
    \centering
    \resizebox{\linewidth}{!}{
    \begin{tabular}{|c|c|c|}
        \hline
        Layer & Input size & Output size\\
        \hline
        \hline
        Conv3x3 + ReLU + MaxPool2x2 & (6, 128, 128) & (32, 64, 64)\\
        \hline
        Conv3x3 + ReLU + MaxPool2x2 & (32, 64, 64) & (64, 32, 32)\\
        \hline
        Conv3x3 + ReLU + MaxPool2x2 & (64, 32, 32) & (64, 16, 16)\\
        \hline
        Conv3x3 + ReLU + MaxPool2x2 & (64, 16, 16) & (128, 8, 8)\\
        \hline
        Conv3x3 + ReLU + MaxPool2x2 & (128, 8, 8) & (128, 4, 4)\\
        \hline
        Conv3x3 + ReLU + MaxPool2x2 & (128, 4, 4) & (256, 2, 2) \\
        \hline
        Flatten & (256, 2, 2) & 1024 \\
        \hline
        Dense + ReLU & 1024 & 128 \\
        \hline
    \end{tabular}
    }
    \caption{MVAE Image encoder}
    \label{tab:MVAEImageEncoder}
\end{table}

\begin{table}[h]
    \centering
    \resizebox{0.7\linewidth}{!}{
    \begin{tabular}{|c|c|c|}
        \hline
        Layer &  Input units & Output units \\
        \hline
        \hline
        Dense+ReLU & 14 & 32 \\
        \hline
        Dense+ReLU & 32 & 64 \\
        \hline
        Dense+ReLU & 64 & 64 \\
        \hline
        Dense+ReLU & 64 & 128 \\
        \hline
        Dense+ReLU & 128 & 128 \\
        \hline
        Dense+ReLU & 128 & 256 \\
        \hline
        Dense+ReLU & 256 & 128 \\
        \hline
    \end{tabular}
    }
    \caption{MVAE Joint encoder}
    \label{tab:MVAEJointEncoder}
\end{table}

\begin{table}[h]
    \centering
    \resizebox{\linewidth}{!}{
    \begin{tabular}{|c|c|c|}
        \hline
        Layer &  Input units & Output units \\
        \hline
        \hline
        \multicolumn{3}{|c|}{Encoder}\\
        \hline
        Concatenate (Image+Joint) & 128, 128 & 256 \\
        \hline
        Dense + Tanh & 256 & 128 mean, 128 std\\
        \hline
        \multicolumn{3}{|c|}{Decoder}\\
        \hline
        Dense+ReLU & 128 & 256 \\
        \hline
        Slice (for image and joint dec.)  & 256 & 128 , 128\\
        \hline
    \end{tabular}
    }
    \caption{MVAE shared encoder-decoder. The activation after the first decoder layer is sliced into two, and each slice is given to a different decoder.}
    \label{tab:MVAEShared}
\end{table}

\begin{table}[h]
    \centering
    \resizebox{\linewidth}{!}{
    \begin{tabular}{|c|c|c|}
        \hline
        Layer & Input size & Output size\\
        \hline
        \hline
        Dense + ReLU & 128 & 1024\\
        \hline
        Reshape & 1024 & (256, 2, 2)\\
        \hline
        Conv3x3 + ReLU + UpSample2x2 & (256, 2, 2) & (256, 4, 4)\\
        \hline
        Conv3x3 + ReLU + UpSample2x2 & (256, 4, 4) & (128, 8, 8)\\
        \hline
        Conv3x3 + ReLU + UpSample2x2 & (128, 8, 8) & (128, 16, 16)\\
        \hline
        Conv3x3 + ReLU + UpSample2x2 & (128, 16, 16) & (64, 32, 32)\\
        \hline
        Conv3x3 + ReLU + UpSample2x2 & (64, 32, 32) & (64, 64, 64)\\
        \hline
        Conv3x3 + ReLU + UpSample2x2 & (64, 64, 64) & (32, 128, 128)\\
        \hline
        Conv3x3 + ReLU & (32, 128, 128) & (16, 128, 128)\\
        \hline
        Conv3x3 + ReLU & (16, 128, 128) & (12, 128, 128)\\
        \hline
        Conv3x3 & (12, 128, 128) & (12, 128, 128)\\
        \hline
    \end{tabular}
    }
    \caption{MVAE Image Decoder. The last activation is sliced into two (6, 128, 128) shaped tensors for mean and std. See the original implementation \cite{zambelli2020multimodal} for further details.}
    \label{tab:MVAEImageDecoder}
\end{table}

\begin{table}[h]
    \centering
    \begin{tabular}{|c|c|c|}
        \hline
        Layer &  Input units & Output units \\
        \hline
        \hline
        Dense+ReLU & 128 & 256 \\
        \hline
        Dense+ReLU & 256 & 128 \\
        \hline
        Dense+ReLU & 128 & 128 \\
        \hline
        Dense+ReLU & 128 & 64 \\
        \hline
        Dense+ReLU & 64 & 64 \\
        \hline
        Dense+ReLU & 64 & 32 \\
        \hline
        Dense & 32 & 28 \\
        \hline
    \end{tabular}
    \caption{MVAE Joint Decoder. The last activation is sliced into two for mean and std.}
    \label{tab:MVAEJointDecoder}
\end{table}

\section{t-SNE Visualization of the Latent Space}

\begin{figure*}[h]
    \centering
    \includegraphics[width=\textwidth]{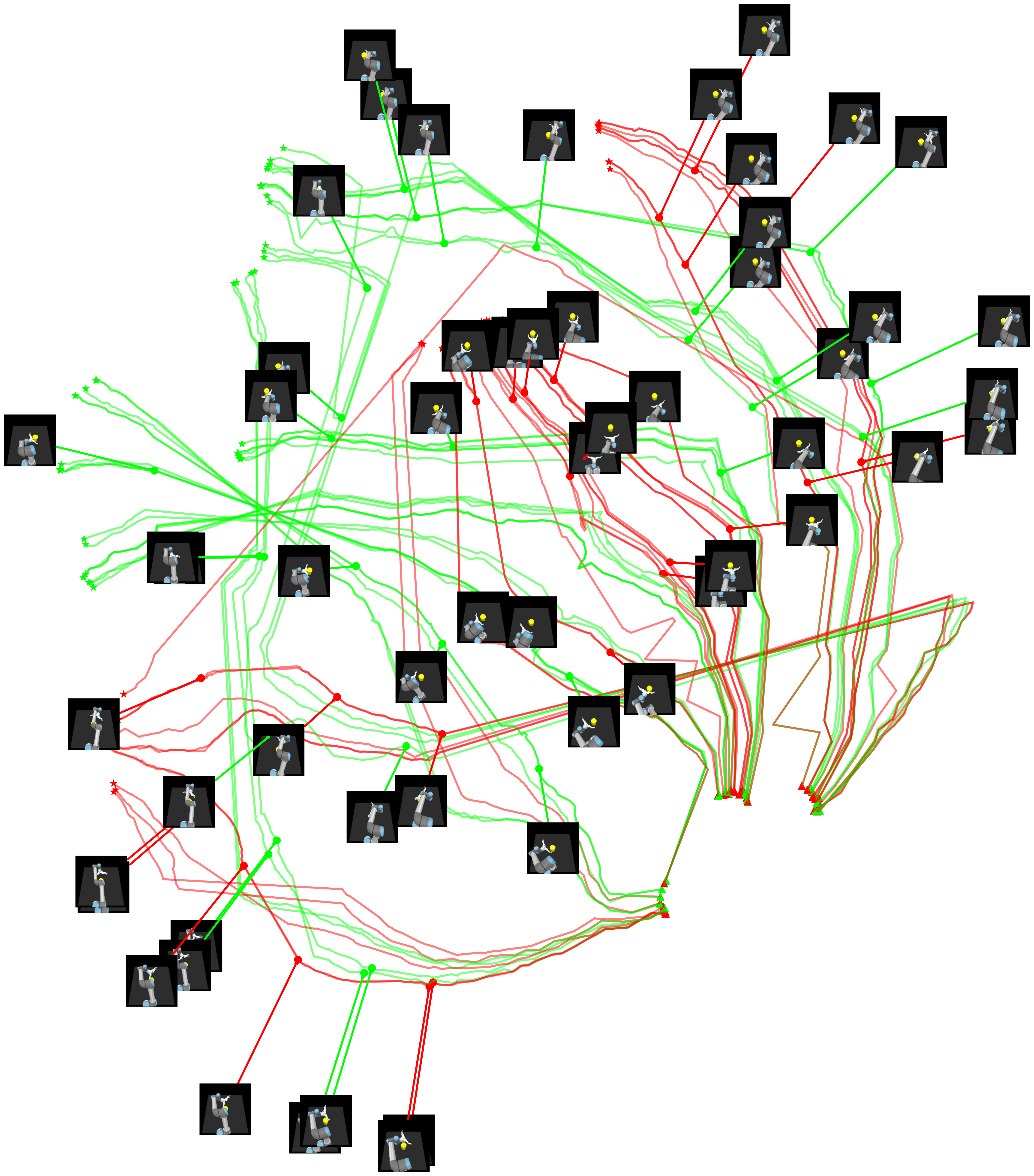}
    \caption{t-SNE \cite{tsne} visualization of the encoder output. Here, green and red represents `move' and `grasp' actions, respectively. The initial and the final point of a trajectory is represented with a triangle and a star, respectively.}
    \label{fig:tsne}
\end{figure*}

In this section, the detailed version of the latent space is investigated. Figure \ref{fig:tsne} shows the encodings of all of the training trajectories in the latent space.

\end{document}